%% file: main.tex
\documentclass[journal]{IEEEtran}
\usepackage{verbatim}
\usepackage{subfiles}
\usepackage{cite}
\usepackage{graphicx}
\usepackage{amsmath}
\usepackage{lipsum}
\ifCLASSOPTIONcompsoc
\usepackage[caption=false, font=normalsize, labelfont=sf, textfont=sf]{subfig}
\else
\usepackage[caption=false, font=footnotesize]{subfig}
\fi
\usepackage{changepage}
\usepackage{float}
\usepackage{algorithm}
\usepackage{algpseudocode}
\usepackage{booktabs}
\usepackage{multirow}
\usepackage{nomencl}
\usepackage{enumitem}
\usepackage{xcolor}
\usepackage{subcaption}
\usepackage[most]{tcolorbox}
\usepackage{amsmath,amssymb,bm}

\makenomenclature
\usepackage{etoolbox}
\usepackage{upgreek}
\usepackage{mathtools}

\usepackage{ifthen}
 \renewcommand{\nomgroup}[1]{%
  \item[\bfseries
  \ifthenelse{\equal{#1}{A}}{Parameters}{%
  \ifthenelse{\equal{#1}{B}}{Variables}{%
  \ifthenelse{\equal{#1}{O}}{Indices and sets}{}}}%
  ]}

\ifCLASSINFOpdf
\else
\fi

\usepackage{color}
\usepackage{soul}
\soulregister\cite7
\soulregister\ref7

\usepackage[dvipsnames]{xcolor}

\begin{document}


\title{From Accounting to Coordination: A Virtual Water-Aware Electricity–Computation–Water Nexus Framework for Data Center Dispatch}
%
%
%

\author{Haiyang You,
Chengwei Lou,
Jin Zhao,~\IEEEmembership {Member, IEEE,}
Yue Zhou,~\IEEEmembership {Member, IEEE,}
Lu Zhang,~\IEEEmembership {Senior Member, IEEE,}
Jin Yang,~\IEEEmembership {Senior Member, IEEE}


}

%
%

\markboth{IEEE Transactions on \LaTeX}%
{SKM: Impact of Generative AI on Regional Carbon Emissions}
%



\maketitle

\begin{abstract}
The expansion of data centers (DCs) drives a sustained increase in electricity demand and associated water withdrawals at generation sites. These withdrawals occur at generation sites and are virtually allocated to demand based on network power flows. Consequently, the actual water footprint of a specific load varies dynamically with generation dispatch and network conditions. Existing approaches typically rely on static statistical accounting to quantify these water footprints. However, these static methods fail to capture how dispatch optimization  and workload relocation dynamically affect water withdrawals. As a result, these static statistical accounting remain decoupled from the optimization process, rendering them incapable of guiding workload relocation or power dispatch to mitigate water stress. To address this limitation, This paper develops an operational electricity–computation–water (ECW) nexus framework that internalizes virtual water impacts directly into power system dispatch. The framework represents dispatch optimization as a differentiable optimization layer embedded within a deep learning architecture, enabling efficient end-to-end learning of coordination policies while preserving operational feasibility. Combined with fixed-point coordination, the framework enforces consistency between virtual water attribution and physical generation side withdrawals. Case studies on the IEEE 30-bus and 118-bus test systems demonstrate reliable convergence, exact power–water consistency, and reductions of approximately 3--5\% in generation-related freshwater withdrawals under water-constrained conditions.
\end{abstract}

\begin{IEEEkeywords}
Virtual water content, Electricity-computation-water nexus, Data center dispatch, End-to-end learning ,Differentiable optimization, Fixed-point coordination
\end{IEEEkeywords}

%
\IEEEpeerreviewmaketitle

\subfile{Nomenclatures}
\subfile{Introduction/1.1}

\section{ECW Nexus System Modeling}
\label{sec:framework}

To enable coordinated dispatch, this section provides the formal mathematical representation of the ECW nexus. We define a composite objective function to internalize water-related impacts into power system decision making. The section then details the physical and operational constraints governing virtual water flows, data center migrations, and network constrained electricity dispatch.

\subfile{MATHEMATICAL_FORMULATION/1.5}

\section{Differentiable Electricity--Computation--Water Nexus Framework}



The optimization problem in Section~\ref{sec:overall_formulation} involves a circular dependency between generation dispatch and virtual water content through the VWC balance~\eqref{eq:vwc_balance}.
This section develops a learning-based computational framework to resolve this dependency across varying operational scenarios.
Section~\ref{sec:constraint_learning} introduces a learning architecture that integrates neural network modules with a differentiable optimization layer. Section~\ref{sec:fixed_point} presents an iterative procedure that coordinates dispatch decisions with virtual water content values.

\subfile{DCOPF_method/1.1}
\subfile{DCOPF_method/1.2}

\section{Case Studies}
This section evaluates the proposed electricity--computation--water 
coordination framework through numerical studies on two test systems.
Section~\ref{sec:ieee30} examines the IEEE 30-bus network to assess 
system-level effects under realistic transmission topology and spatially 
heterogeneous water scarcity, with particular attention to dispatch 
patterns, virtual water redistribution, and sensitivity to key parameters. Section~\ref{sec:ieee118} demonstrates scalability of the framework on the IEEE 118-bus system. Sections~\ref{sec:validation} and~\ref{sec:vl_coordination} then consider a 5-bus system to examine convergence behavior of the fixed-point iteration  and to verify consistency between virtual and physical water accounting.

\subfile{Case_study/1.1}

\subfile{Case_study/1.2}

\subfile{Case_study/1.3}
\section{Conclusion} 
This paper develops a learning-based coordination framework that embeds dispatch optimization as a differentiable layer, enabling virtual water considerations to directly inform electricity dispatch decisions. By treating virtual water content as an endogenous outcome of dispatch and power flows, the framework allows water scarcity considerations to inform 
system operation through end-to-end learning rather than retrospective statistical accounting. Numerical studies on IEEE 30-bus and 118-bus systems show that coordinated workload relocation reduces generation-related water withdrawals by 3--5\% while preserving power balance and network feasibility. Flexible computing demand shifts electricity consumption toward regions with lower water-intensity generation, thereby alleviating localized water stress without changing aggregate system load. Sensitivity analysis further indicates that the effectiveness of coordination depends on water availability signals and network constraints, with diminishing returns once transmission and generation limits become binding. At the algorithmic level, the fixed-point procedure guarantees consistency between virtual and physical water accounting. Validation on a 5-bus system confirms that virtual water at consumption nodes matches physical withdrawals at generation sites within numerical tolerance. Overall, the proposed framework provides a tractable learning-based approach for integrating virtual water dynamics into power system operations and offers a foundation for coordinated ECW management in systems with growing data center penetration.











%

\subfile{appendix}




\ifCLASSOPTIONcaptionsoff
  \newpage
\fi

\bibliographystyle{IEEEtran}
\bibliography{REF}

\end{document}

%% file: Introduction/1.1.tex
\section{Introduction}

The ongoing process of digitalization is reshaping modern societies through the widespread adoption of smart devices, cloud-based services, and globally interconnected computing infrastructures \cite{raihan2024review,bocean2023eu}. This transformation has given rise to a rapidly expanding global digital economy characterized by massive data generation and processing demand. The recent emergence of artificial intelligence (AI), particularly large language models (LLMs), has intensified this trajectory due to their enormous computational requirements\cite{ruan2024applying}. Meeting these demands necessitates the deployment of large-scale data centers (DCs) on an unprecedented scale. According to the International Energy Agency (IEA), global DC electricity demand is set to more than double to around 945~TWh by 2030~\cite{iea2025energy}. This demand is growing at nearly 12\% annually and already accounts for approximately 1.5\% of global electricity consumption~\cite{nature2025data, bashir2025explained, devries2023growing,farfan2023gone}. While the electricity implications of this digital growth are widely studied, its impact on water resources has recently attracted systematic attention. 

\begin{figure}[ht]
  \centering
  \includegraphics[width=\linewidth]{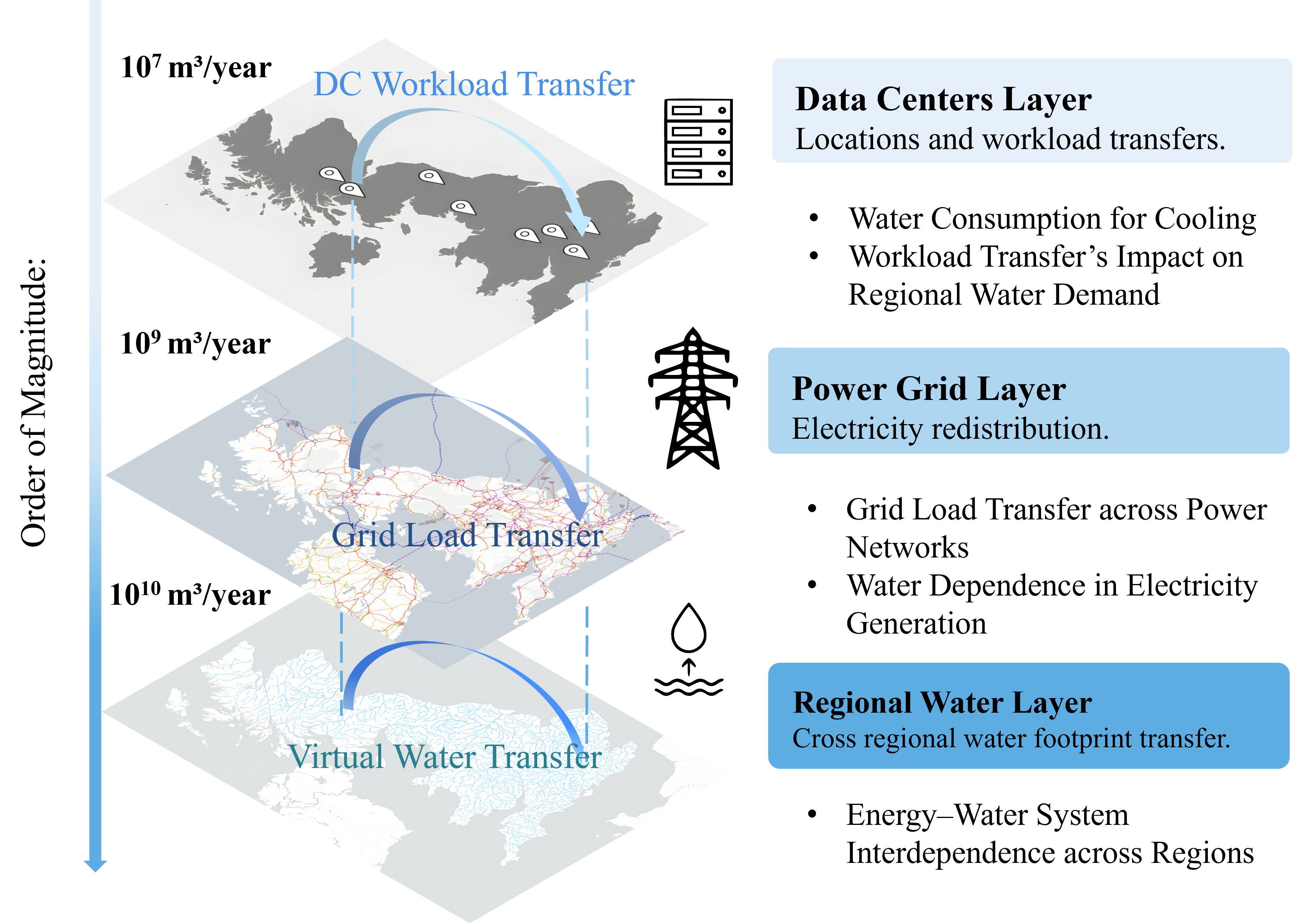}
  \caption{ECW nexus framework.}
  \label{fig:dc-grid-water}
\end{figure}

Beyond direct cooling requirements, DC operations involve water withdrawals embedded in power system operation. The expansion of DC electricity demand driven by these digital trends has led to a substantial increase in water withdrawals\cite{yates2024modeling}. Global DC operations currently withdraw up to 1.7 billion liters of water per day including direct on-site usage and indirect water consumption~\cite{mytton2021data,lei2025water,siddik2024spatially}. However, understanding the full scope of this water impact requires moving beyond direct consumption metrics. The relationship between DC operations and water resources operates through multiple interconnected layers, necessitating a modeling framework that reflects both grid physics and spatial water dynamics, and that can quantify their impacts on regional water withdrawal patterns. Characterizing the interdependence between power system operation and water redistribution requires a modeling framework that reflects both grid physics and spatial water dynamics. As illustrated in Fig.~\ref{fig:dc-grid-water}, these interactions manifest as a multi-layer electricity--computation--water (ECW) nexus. At the top layer, computing workload transfers between data centers alter regional electricity demand. These changes trigger cross regional power exchanges within the middle-layer transmission network. The resulting adjustments in generation dispatch reshape water withdrawal patterns in the bottom layer and redistribute the virtual water embodied in power flows~\cite{zhang2020flexibility}. The hierarchy in Fig.~\ref{fig:dc-grid-water} reveals that water-related impacts exhibit a progressive scaling effect. Associated water footprint estimates expand from the direct footprint of DCs ($10^7$~m$^3$/year) to the broader redistribution enabled by the power grid ($10^9$~m$^3$/year) and finally to the regional water footprint driven by large scale withdrawals ($10^{10}$~m$^3$/year). This process links digital operations with physical resource flows and highlights the necessity of integrated ECW modeling.

Quantifying these grid mediated water impacts requires a framework that can trace water withdrawals from generation sites to end use consumption. Prior studies utilize the virtual water framework to characterize these embedded impacts. Virtual water represents water resources embedded during electricity generation and transferred across the power grid from production sites to consumption regions~\cite{chini2018virtual,hogade2022survey, zhao2020water,chini2020changing}. Within this framework, the power grid serves as a physical medium to transfer water-related environmental burdens from generation sites to end-use consumption points~\cite{tian2022water,liao2018categorising,jin2025virtual}. Existing research applies multi-regional input-output models~\cite{liao2018virtual,guo2016network,lin2019provincial} and node flow formulations~\cite{zhu2015embodiment,zhang2017node,liu2022optimal,wan2025real,alnahhal2023water} to quantify these transfers~\cite{zhang2021tracking,wang2021virtual,jin2025virtual}. However, these accounting-based methods are inherently retrospective because VWC intensities are calculated only after generation schedules and power flows are realized. Consequently, these metrics remain decoupled from real time operational decision making~\cite{li2023making,iea4e2025datacentre}. This structural separation creates a fundamental limitation where data centers possess significant spatial flexibility but dispatch frameworks do not incorporate water-related signals into load allocation. Virtual water analysis remains a descriptive tool and cannot guide proactive optimization decisions during system operations.

To bridge this gap between retrospective accounting and operational coordination, establishing explicit coupling between virtual water metrics and dispatch decision-making becomes essential. Conventional approaches calculate virtual water intensities after dispatch solutions are realized, treating these as independent sequential processes. The power system first optimizes electricity flows to minimize generation costs, and water metrics are subsequently computed based on the resulting generation mix. This separation prevents water-related signals from influencing workload allocation because the intensities used to guide decisions remain fixed at pre-calculated values. However, any reallocation of data center workloads alters regional electricity demand patterns, triggering adjustments in cross regional power exchanges and generation dispatch. These changes reshape the spatial distribution of water withdrawals and modify the virtual water content embodied in power flows. The fundamental challenge is that water intensities depend on dispatch outcomes, while optimal dispatch should ideally reflect water considerations. This creates a circular dependency: water intensities depend on dispatch outcomes, while water-aware dispatch requires knowing these intensities in advance. Resolving this circular dependency necessitates a framework where dispatch optimization and virtual water determination evolve jointly rather than statically.

To resolve this circular dependency, this study develops a learning based framework that embeds dispatch optimization as a differentiable optimization layer within an end-to-end learning architecture~\cite{amos2017optnet,liang2024joint}. Through the implicit function theorem~\cite{krantz2002implicit}, gradients can be computed and propagated through this differentiable layer. The neural network outputs are projected onto the feasible set, and physical constraints are preserved~\cite{11230858,donti2021dc3,agrawal2019differentiable,%
li2023learning}. This enables the joint evolution of dispatch decisions and virtual water intensities. These gradients establish computational pathways through which changes in DC workload distribution propagate across the transmission network to affect regional water withdrawal patterns. The resulting framework allows water-related signals to actively guide coordination decisions through gradient-based learning, shifting 
computational burden from real-time operations to offline training. Building on this differentiable structure, a fixed-point coordination procedure is introduced to iteratively reconcile dispatch solutions with virtual water intensities~\cite{safdarian2018time}. Each iteration updates dispatch decisions based on current water signals and recalculates intensities based on realized generation patterns. This iterative process continues until electricity flows and water withdrawals reach mutual consistency. The resulting framework enables virtual water metrics to transition from retrospective accounting tools to operational coordination signals that actively inform workload allocation decisions.

The main contributions of this study are summarized below. \begin{itemize} 
\item A coordination framework is developed within the ECW nexus that integrates DC workload scheduling with virtual water aware dispatch. This framework enables water-related environmental impacts to inform operational decisions rather than serve solely as retrospective metrics. 
\item A coordination framework is developed that embeds dispatch optimization as a differentiable layer within a learning architecture. Using the implicit function theorem, gradients can be computed and propagated through this layer, enabling end-to-end learning of coordination policies and the joint evolution of dispatch decisions and virtual water intensities, while establishing direct computational pathways from workload allocation to regional water footprints.
\item A fixed-point coordination algorithm is introduced to resolve the circular dependency between dispatch outcomes and virtual water intensities. This iterative procedure reconciles electricity flows with water withdrawal patterns until convergence is achieved.
\end{itemize}

Numerical experiments on representative test systems demonstrate that the proposed framework with differentiable optimization layers and fixed-point coordination reliably resolves the circular dependency between virtual water intensities and dispatch outcomes while maintaining exact consistency between virtual and physical water accounting. Unlike conventional decoupled approaches, the framework enables water-related signals to actively guide workload allocation and generation scheduling through gradient-based learning. The resulting model provides a scalable and operationally grounded foundation for 
incorporating water-related considerations into power system 
operation under the continued expansion of digital 
infrastructure.

%% file: MATHEMATICAL_FORMULATION/1.5.tex
\subsection{Overall Optimization Formulation}
\label{sec:overall_formulation}

An optimization framework is developed to coordinate power system dispatch, spatial workload allocation, and water scarcity considerations within a unified decision making setting. The formulation seeks generation schedules and workload allocations that balance economic efficiency, water sustainability, and operational stability. This objective is realized by minimizing a composite cost function that accounts for electricity generation costs, virtual water impacts, and workload migration penalties:
\begin{equation}
\min 
C^{\mathrm{gen}}
+
C^{\mathrm{water}}
+
C^{\mathrm{reg}},
\label{eq:overall_objective_compact}
\end{equation}
where the three components are defined as:
\begin{subequations}
\label{eq:overall_cost_terms}
\begin{align}
C^{\mathrm{gen}}
&=
\sum_{g \in \mathcal{G}} C_g(P_g),
\label{eq:cost_gen}
\\
C^{\mathrm{water}}
&=
\lambda_w
\sum_{n \in \mathcal{N}}
\mathrm{VWC}_n\, \widetilde{P}_n,
\label{eq:cost_water}
\\
C^{\mathrm{reg}}
&=
\frac{\rho^{\mathrm{mig}}}{2}
\sum_{d \in \mathcal{D}} \sum_{n \in \mathcal{N}}
\left(\widehat{w}_{d,n}-w_{d,n}\right)^2 .
\label{eq:cost_reg}
\end{align}
\end{subequations}
The term $C^{\mathrm{gen}}$ captures the operating cost of conventional electricity generation.  The term $C^{\mathrm{water}}$ internalizes the water-related environmental impact by weighting the effective nodal electricity demand $\widetilde{P}_n$ with the corresponding nodal virtual water content $\mathrm{VWC}_n$. The regularization term $C^{\mathrm{reg}}$ quantifies the cost of workload reallocation by penalizing deviations of the realized workload allocation $\widehat{w}_{d,n}$ from the baseline allocation $w_{d,n}$.

In addition to the objective function, explicit constraints are imposed to represent limits on physical water withdrawals. The water withdrawal for electricity generation at node $n$ is defined as
\begin{equation}
W^{\mathrm{gen}}_{n}
=
\sum_{g \in \mathcal{G}_n} \kappa_g \, P_g,
\qquad \forall n \in \mathcal{N}.
\label{eq:node_gen_water}
\end{equation}
where $\kappa_g$ is the water withdrawal coefficient of generator $g$. To reflect system-level water scarcity, a scarcity-weighted water budget is enforced:
\begin{equation}
\sum_{n \in \mathcal{N}} s_n \, W^{\mathrm{gen}}_{n}
\le
W^{\mathrm{budget}},
\label{eq:scarcity_weighted_budget}
\end{equation}
where $s_n$ denotes the water stress coefficient at node $n$ and $W^{\mathrm{budget}}$ is the total allowable withdrawal.

The optimization is further constrained by the models described in the following subsections. These include the virtual water balance in Eq.~\eqref{eq:vwc_balance}, the DC workload allocation and Quality of Service constraints in Eq.~\eqref{eq:w_colsum}--\eqref{eq:latency_constraint}, and the power system operational constraints in Eq.~\eqref{eq:power_system_optimization}.

This formulation exhibits a circular dependency: $C^{\mathrm{water}}$ depends on $\mathrm{VWC}_n$, which is determined by dispatch outcomes through Eq.~\eqref{eq:vwc_balance}. Section~\ref{sec:constraint_learning} develops an iterative procedure to resolve this coupling.
\subsection{Virtual Water Content of Electricity Consumption}
\label{sec:vwc_model}

The nodal $\mathrm{VWC}_n$ represents the average water withdrawal intensity of electricity demand at bus $n$, calculated by dividing the total virtual water inflow by the local electricity consumption. Under the proportional sharing principle~\cite{bialek2004proportional}, the virtual water content of incoming flows is traced back to upstream generation sources, with each node's VWC reflecting the weighted average of its supply mix. Accordingly, the nodal virtual water balance is written as

\begin{equation}
\mathrm{VWC}_n
=
\frac{
\sum_{g \in \mathcal{G}_n} \kappa_g \, P_g
+
\sum_{m:(m,n)\in\mathcal{L}} \mathrm{VWC}_m\,[f_{mn}]^{+}
}{
\widetilde{P}_n
},
\label{eq:vwc_balance}
\end{equation}
where $[f_{mn}]^{+}=\max(f_{mn},0)$ represents incoming power flow from bus $m$ to bus $n$. The denominator $\widetilde{P}_n$ is the effective nodal electricity demand, including both baseline load and DC-induced load, and therefore varies with workload allocation decisions.

Equation~\eqref{eq:vwc_balance} shows that virtual water content depends on generation dispatch and network power flows. At the same time, generation and flows are determined by minimizing an objective function that contains $\mathrm{VWC}_n$. This mutual dependence between virtual water content and operational decisions creates a circular coupling that cannot be resolved through standard single-stage optimization, motivating the iterative coordination framework developed in Section~\ref{sec:constraint_learning}.

\subsection{DC Workload Integration}
\label{sec:dc_workload}

DCs are connected to a subset of transmission buses $d \in \mathcal{D} \subseteq \mathcal{N}$ and are treated as electricity intensive loads. Each bus $n$ generates computing workload $\delta_n$, expressed in electricity equivalent units, part of which can be spatially redistributed among DCs.

Let $w_{d,n}$ denote the portion of workload from bus $n$ that is initially assigned to DC $d$. The baseline allocation satisfies
\begin{align}
\sum_{d \in \mathcal{D}} w_{d,n} &= \delta_n,
\qquad \forall n \in \mathcal{N},
\label{eq:w_colsum} \\
\sum_{n \in \mathcal{N}} w_{d,n} &= \vartheta_d,
\qquad \forall d \in \mathcal{D},
\label{eq:w_rowsum}
\end{align}
where $\vartheta_d$ is the initial total workload assigned to DC $d$.

Workload migration is modeled using a virtual transfer network with flow vector $\phi$ and incidence matrix $A^{\mathrm{DC}}$. The realized workload $\widehat{\vartheta}_d$ must satisfy
\begin{equation}
A^{\mathrm{DC}} \phi = \widehat{\vartheta} - \vartheta,
\label{eq:dc_flow_balance}
\end{equation}
with capacity limits
\begin{equation}
- \overline{\Phi}_k \le \phi_k \le \overline{\Phi}_k,
\qquad \forall k \in \mathcal{K}^{\mathrm{DC}}.
\label{eq:dc_transfer_capacity}
\end{equation}

To describe the composition of migrated workload, let $\widehat{w}_{d,n}$ denote the portion of workload from bus $n$ that is eventually processed at DC $d$. It satisfies
\begin{align}
\sum_{n \in \mathcal{N}} \widehat{w}_{d,n} &= \widehat{\vartheta}_d,
\qquad \forall d \in \mathcal{D}, 
\label{eq:what_rowsum} \\
\sum_{d \in \mathcal{D}} \widehat{w}_{d,n} &= \delta_n,
\qquad \forall n \in \mathcal{N}, 
\label{eq:what_colsum} \\
\widehat{w}_{d,n} &\ge 0,
\qquad \forall d \in \mathcal{D},\ n \in \mathcal{N}.
\label{eq:what_nonneg}
\end{align}
Consistency with aggregate migration is enforced by
\begin{equation}
\sum_{n \in \mathcal{N}} \widehat{w}_{d,n}
-
\sum_{n \in \mathcal{N}} w_{d,n}
=
\big[A^{\mathrm{DC}}\phi\big]_d,
\qquad \forall d \in \mathcal{D}.
\label{eq:w_phi_coupling}
\end{equation}

To limit service degradation, a latency-based Quality of Service constraint is imposed. The latency coefficient $C_{d,n}$ is defined based on geographical distance between regions, and $L^{\mathrm{nom}}$ represents the baseline latency under initial allocation. The realized latency
\begin{equation}
L = \sum_{d \in \mathcal{D}} \sum_{n \in \mathcal{N}} C_{d,n}\,\widehat{w}_{d,n}
\label{eq:realized_latency}
\end{equation}
must satisfy
\begin{equation}
L \le (1+\alpha^{\mathrm{QoS}})\, L^{\mathrm{nom}}.
\label{eq:latency_constraint}
\end{equation}

Each DC converts workload into electricity demand using coefficient $\Gamma_d$, so that the DC-induced nodal load is
\begin{equation}
P^{\mathrm{DC}}_n
=
\sum_{d \in \mathcal{D}_n} \Gamma_d \, \widehat{\vartheta}_d,
\qquad \forall n \in \mathcal{N}.
\label{eq:dc_power_demand}
\end{equation}

\subsection{Power System Operational Constraints}
\label{sec:power_system}

The power system is modeled as a steady-state network-constrained dispatch problem. The total nodal electricity demand is
\begin{equation}
\widetilde{P}_n = P_n^{\mathrm{base}} + P_n^{\mathrm{DC}},
\qquad \forall n \in \mathcal{N}.
\label{eq:net_demand}
\end{equation}

The operational constraints are
\begin{subequations}
\label{eq:power_system_optimization}
\begin{align}
\sum_{g \in \mathcal{G}_n} P_g
-
\widetilde{P}_n
&=
\sum_{m \in \Omega_n}
b_{nm}(\theta_n-\theta_m),
\label{eq:ps_balance}
\\
- \bar{F}_{nm}
\le
b_{nm}(\theta_n-\theta_m)
&\le
\bar{F}_{nm},
\qquad \forall (n,m) \in \mathcal{L},
\label{eq:ps_line}
\\
\underline{P}_g \le P_g &\le \overline{P}_g,
\qquad \forall g \in \mathcal{G},
\label{eq:ps_gen}
\\
\theta_{n_0} &= 0.
\label{eq:ps_ref}
\end{align}
\end{subequations}

Here $P_g$ is the output of generator $g$, $\theta_n$ is the voltage angle at bus $n$, $b_{nm}$ is the susceptance of line $(n,m)$, and $\bar{F}_{nm}$ is its capacity limit. Bus $n_0$ is the reference bus.

%% file: DCOPF_method/1.1.tex
\subsection{Differentiable Optimization Layer}
\label{sec:constraint_learning}

The optimization problem formulated in Section~\ref{sec:overall_formulation}
exhibits a circular dependency through the VWC balance~\eqref{eq:vwc_balance}: 
nodal VWC values depend on generation dispatch and power flows, while 
dispatch decisions are influenced by VWC through the water-related cost 
term~\eqref{eq:cost_water}.
Conventional approaches compute dispatch first, then calculate VWC as a 
post-hoc metric.
This sequential treatment prevents water-related signals from guiding 
dispatch decisions, as VWC values remain fixed during optimization and 
do not reflect how workload reallocation would alter generation patterns.

To resolve this circular dependency, the dispatch optimization is embedded as a differentiable layer within the learning-based coordination framework. Through the implicit function theorem, gradients of the operational 
objective backpropagate through the dispatch constraints.
This enables workload allocation decisions to account for their downstream 
effects on generation dispatch, power flows, and water withdrawals.

The framework adopts a learning architecture that integrates neural network modules with a differentiable optimization layer as illustrated in
Fig.~\ref{fig:flowchart}.
A multi layer perceptron $f_{\boldsymbol{\psi}}$ predicts initial 
generator dispatch from effective nodal demand:
\begin{equation}
\widehat{P}_g = f_{\boldsymbol{\psi}}(\widetilde{P}).
\label{eq:nn_pred}
\end{equation}
This prediction accelerates convergence across repeated scenarios but may violate operational constraints. Feasibility is enforced by projecting $\widehat{P}_g$ onto the constraint set through a differentiable optimization layer:


\begin{figure*}[h!]
    \centering
    \includegraphics[width=0.8\linewidth]{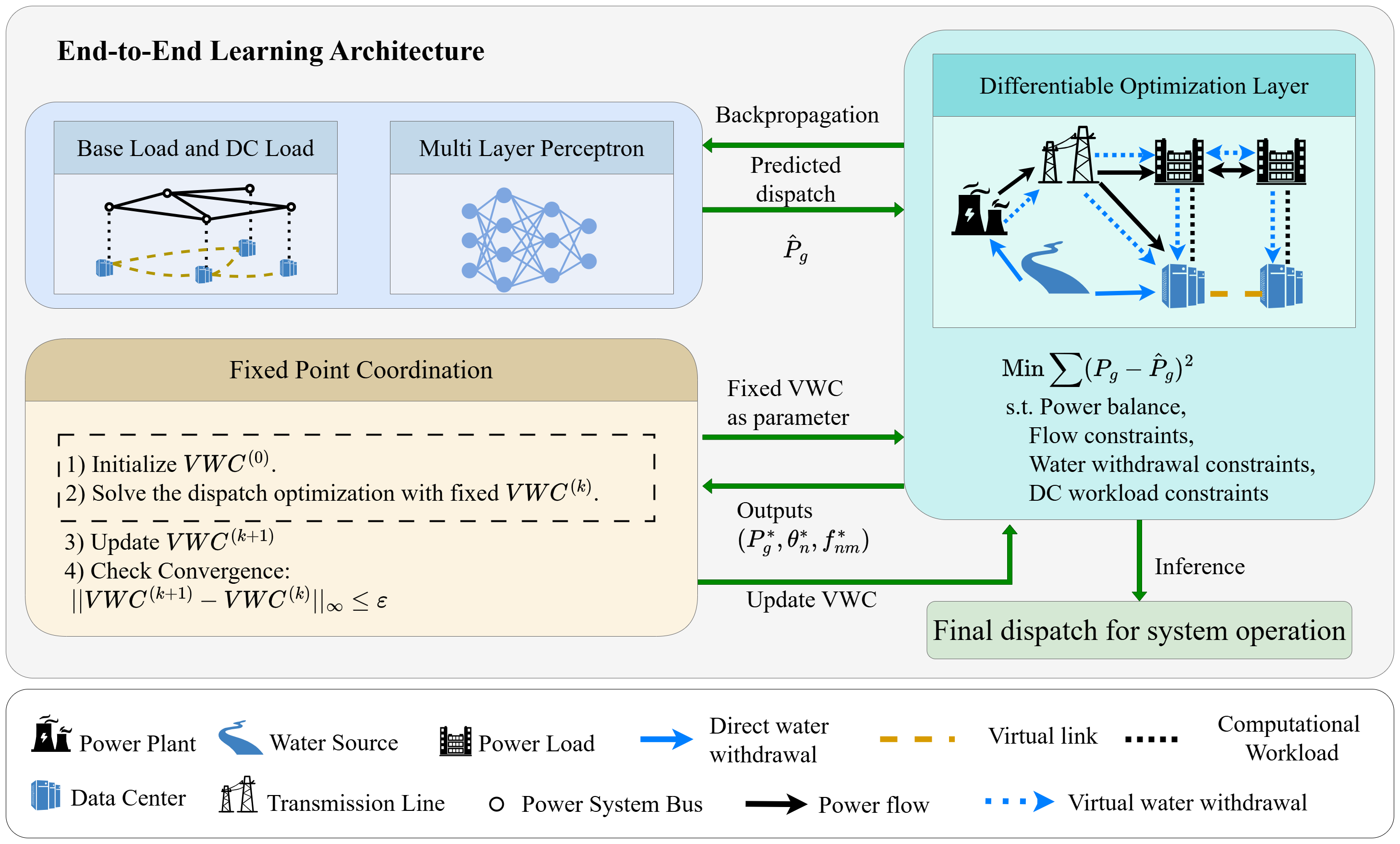}
    \caption{Overview of the coordination framework with differentiable 
    optimization layers and fixed-point iteration.}
    \label{fig:flowchart}
\end{figure*}

\begin{equation}
\begin{aligned}
\min \quad &
\sum_{g \in \mathcal{G}}
\left(P_g-\widehat{P}_g\right)^2
\\
\text{s.t.}\quad &
\text{Constraints~\eqref{eq:ps_balance}--\eqref{eq:ps_ref},
\eqref{eq:w_colsum}--\eqref{eq:w_rowsum},}
\\
&
\text{\eqref{eq:dc_flow_balance}--\eqref{eq:dc_transfer_capacity},
\eqref{eq:what_rowsum}--\eqref{eq:what_nonneg},}
\\
&
\text{\eqref{eq:w_phi_coupling},
\eqref{eq:latency_constraint},
\eqref{eq:vwc_balance},
\eqref{eq:scarcity_weighted_budget}.}
\end{aligned}
\label{eq:cvx_layer_consistent}
\end{equation}
In this correction step, $\mathrm{VWC}_n$ is treated as fixed and provided 
by the iterative procedure in Section~\ref{sec:fixed_point}.
Under fixed $\mathrm{VWC}_n$, problem~\eqref{eq:cvx_layer_consistent} 
defines a convex program that can be efficiently solved and differentiated.

The optimization layer is implemented as a differentiable convex program~\cite{agrawal2019cvxpylayer}. 
Gradients are obtained by implicit differentiation of the KKT conditions. 
Appendix~\ref{app:gradient_derivation} provides the derivation for the QP instance used in our formulation.

These gradients reveal how changes in workload allocation propagate through 
dispatch decisions to affect water withdrawals.
Specifically, a change in $\widehat{w}_{d,n}$ alters effective demand 
$\widetilde{P}_n$, triggering adjustments in generation dispatch $P_g$ 
and power flows $f_{nm}$, which in turn modify the water-related cost 
$C^{\mathrm{water}}$ through~\eqref{eq:cost_water}.

Training minimizes the loss function:
\begin{equation}
\begin{aligned}
\mathcal{L}(\boldsymbol{\psi})
=
\mathbb{E}\!\left[
C^{\mathrm{gen}}
+
C^{\mathrm{water}}
+
C^{\mathrm{reg}}
\right]
+
\lambda_{\mathrm{acc}}
\mathbb{E}\!\left[
\left\|
\widehat{P}_g
-
P_g^\star
\right\|_2^2
\right],
\end{aligned}
\label{eq:training_loss_consistent}
\end{equation}
where $P_g^\star$ is the feasible dispatch from~\eqref{eq:cvx_layer_consistent}.
The first term captures operational costs including generation, water 
impacts, and workload migration penalties.
The second term with weight $\lambda_{\mathrm{acc}} > 0$ encourages 
accurate predictions.
Gradients of $C^{\mathrm{water}}$ flow through the differentiable layer, 
allowing the predictor to learn policies that reduce water withdrawals 
by directing computing demand toward regions with lower-intensity generation.

The differentiable architecture shifts VWC from a retrospective accounting 
tool to an operational signal.
After each correction step, the VWC balance~\eqref{eq:vwc_balance} is 
evaluated to update $\mathrm{VWC}_n$, which then influences the next 
iteration through~\eqref{eq:cost_water}.
This iterative process, detailed in Section~\ref{sec:fixed_point}, 
reconciles dispatch decisions with virtual water intensities until 
convergence is achieved.

%% file: DCOPF_method/1.2.tex
\subsection{Iterative Coordination for Endogenous Virtual Water Content}
\label{sec:fixed_point}

This architecture introduced in
Section~\ref{sec:constraint_learning} treats nodal virtual water content
values $\mathrm{VWC}_n$ as fixed parameters when solving the
correction problem~\eqref{eq:cvx_layer_consistent}.
This treatment enables the differentiable layer to enforce all operational constraints while maintaining differentiability.
However, the virtual water content is not an exogenous system parameter but
an endogenous quantity that must be determined consistently with the
realized generation dispatch and power flow patterns.

As defined in~\eqref{eq:vwc_balance}, the virtual water content at each bus
depends on the generation dispatch $P_g$, the power flows $f_{nm}$, and the
effective nodal demand $\widetilde{P}_n$.
At the same time, virtual water content values enter the operational
objective through the water-related cost term~\eqref{eq:cost_water}, thereby
influencing dispatch decisions.
This circular dependence necessitates an iterative procedure in which 
dispatch decisions and nodal virtual water content values are updated 
sequentially until mutual consistency is achieved.

Given an estimate
$\mathrm{VWC}^{(k)}=\{\mathrm{VWC}_n^{(k)}\}_{n\in\mathcal{N}}$ at iteration $k$,
the correction problem~\eqref{eq:cvx_layer_consistent} is solved with
$\mathrm{VWC}^{(k)}$ treated as fixed, yielding a dispatch and voltage angle
solution $\{P_g^{(k)}, \theta_n^{(k)}\}$.
The corresponding power flows are computed from the voltage angles as
$f_{nm}^{(k)} = b_{nm}(\theta_n^{(k)} - \theta_m^{(k)})$.
The nodal virtual water content values are then updated by evaluating the
virtual water balance~\eqref{eq:vwc_balance} under the resulting dispatch
and flows, producing a revised estimate $\mathrm{VWC}^{(k+1)}$.

This procedure defines a fixed-point iteration
\begin{equation}
\mathrm{VWC}^{(k+1)} = T\!\left(\mathrm{VWC}^{(k)}\right),
\label{eq:fp_mapping_consistent}
\end{equation}
where the operator $T$ implicitly encapsulates solving
\eqref{eq:cvx_layer_consistent} followed by evaluating
\eqref{eq:vwc_balance}.
A fixed point of this mapping corresponds to a solution in which the
virtual water content values used in the dispatch optimization are
consistent with those implied by the resulting generation and power flow
patterns.

To improve numerical stability, a damped update scheme is applied,
\begin{equation}
\begin{aligned}
\mathrm{VWC}^{(k+1)}
&=
\alpha\,T\!\left(\mathrm{VWC}^{(k)}\right) \\
&\quad + (1-\alpha)\,\mathrm{VWC}^{(k)},
\qquad \alpha \in (0,1),
\end{aligned}
\label{eq:damped_fp}
\end{equation}
which moderates successive updates.
The iteration terminates when
\begin{equation}
\left\|\mathrm{VWC}^{(k+1)} - \mathrm{VWC}^{(k)}\right\|_\infty < \epsilon,
\label{eq:convergence_criterion}
\end{equation}
where $\epsilon > 0$ is a prescribed tolerance.

%% file: Case_study/1.1.tex
\subsection{Analysis on the IEEE 30-Bus System}
\label{sec:ieee30}

The proposed framework is evaluated on the IEEE 30-bus system, which
includes six generators and heterogeneous nodal loads.
Generator water withdrawal coefficients are specified according to
regional technology characteristics and range from
2.2 to 2.6~m$^3$/MWh.
The scarcity-weighted system-level water constraint
\eqref{eq:scarcity_weighted_budget} is enforced with
$W^{\mathrm{budget}}$ set to 9000~m$^3$/h.
The capacity of each virtual workload migration link is set to
$\overline{\Phi}_k = 20$~MW (electricity-load equivalent).
Transmission limits follow the standard IEEE 30-bus configuration.

\begin{figure}[h!]
    \centering
    \includegraphics[width=\columnwidth]{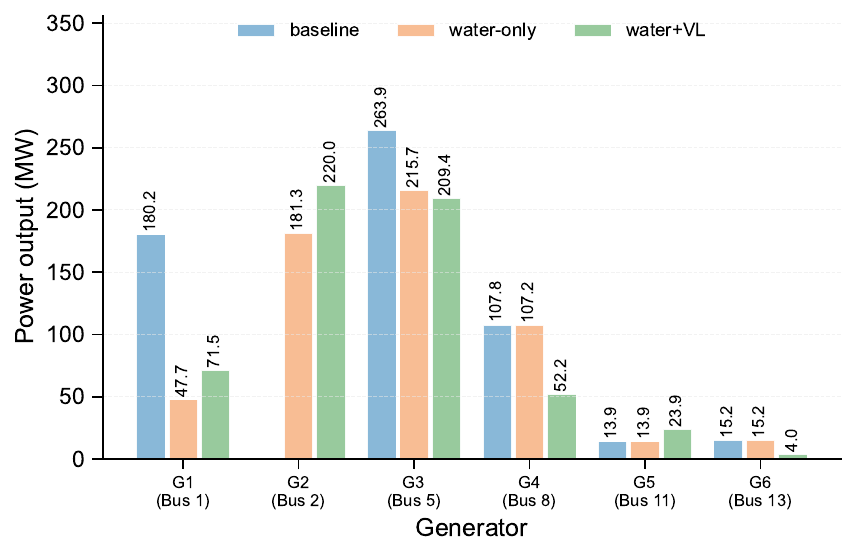}
    \caption{Generator dispatch across baseline, water-only, and Water+VL configurations.}
    \label{fig:pg_30bus}
\end{figure}

Fig.~\ref{fig:pg_30bus} reports generator dispatch under three
configurations: a baseline case without water considerations, a
water-only case with water-related constraints but no workload
migration, and a coordinated case with both water constraints and
virtual links (Water+VL).
In the baseline case, generation is concentrated at large units located
at Buses~1 and~5, while smaller generators at Buses~11 and~13 provide
limited output.
Introducing the system-level water constraint shifts dispatch toward
lower-intensity generators.
Allowing workload migration further modifies the dispatch pattern:
output from the generator at Bus~5 decreases from 263.9~MW to 209.4~MW,
while generation at Bus~2 increases to 220.0~MW.

\begin{table}[h]
\centering
\caption{System performance and workload transfers under Water+VL configuration on the IEEE 30-bus system.}
\label{tab:ieee30_results}
\setlength{\tabcolsep}{4pt}
\footnotesize
\begin{tabular}{lccc}
\toprule
\multicolumn{4}{l}{System-level generation and water withdrawal} \\
\midrule
Case & Generation (MW) & Water withdrawal (m$^3$/h) & Change (\%) \\
\midrule
Baseline & 580.97 & 1368.21 & -- \\
Water+VL & 580.97 & 1309.16 & $-4.3$ \\
\midrule
\multicolumn{4}{c}{} \\
\midrule
\multicolumn{4}{l}{Workload transfers under Water+VL} \\
\midrule
Origin Bus & To Bus 1 (MW) & To Bus 2 (MW) & Total (MW) \\
\midrule
Bus 8  & 7.6 & 8.4 & 16.0 \\
Bus 23 & 5.3 & 4.8 & 10.1 \\
Bus 27 & 4.2 & 4.6 &  8.8 \\
Bus 30 & 2.4 & 2.6 &  5.0 \\
\midrule
Total received & 19.5 & 20.4 & 39.9 \\
\bottomrule
\end{tabular}
\end{table}

Table~\ref{tab:ieee30_results} summarizes aggregate system quantities and workload migration patterns under the Water+VL configuration. Total electricity generation remains unchanged, while coordinated operation reduces generation-related water withdrawals by 4.3\%. The reconciliation between virtual water consumption and physical water withdrawals is preserved, with $\sum_{n} \mathrm{VWC}_{n}\,\widetilde{P}_{n} = \sum_{n} W^{\mathrm{gen}}_{n}$, where $W^{\mathrm{gen}}_{n}$ is defined in~\eqref{eq:node_gen_water}. Workloads are primarily transferred from Buses 8, 23, 27, and 30 toward Buses 1 and 2. Bus 8 exhibits the largest transfers, with 8.4 MW directed to Bus 2 and 7.6 MW to Bus 1. In total, Bus 2 receives 20.4 MW while Bus 1 receives 19.5 MW. This asymmetric pattern reflects the influence of the virtual water weighted cost term in~\eqref{eq:cost_water}, which incentivizes allocation of flexible computing demand toward regions supplied by lower-intensity generation.

\begin{figure}[h!]
    \centering
    \includegraphics[width=\columnwidth]{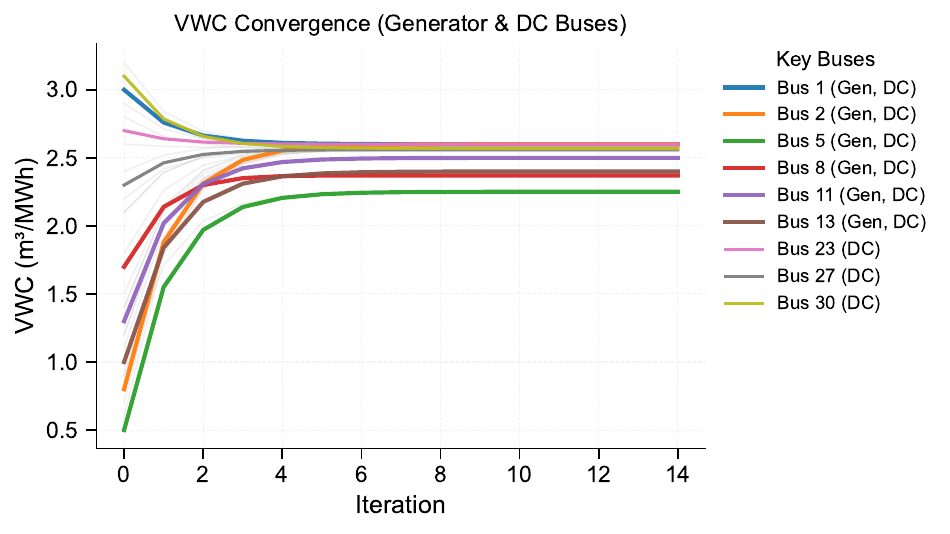}
    \caption{Convergence of virtual water content during fixed-point iteration.}
    \label{fig:vwc_conv_30bus}
\end{figure}

The fixed-point iteration converges within 10--12 iterations.
Across all buses, the converged virtual water content lies between
2.3 and 2.7~m$^3$/MWh, consistent with the range of generator water
intensities.
Nodes with local generation exhibit VWC values close to their
corresponding $\kappa_g$, while demand-dominated nodes reflect blended
contributions from multiple supply sources through transmission and
workload migration.

\begin{figure}[!t]
    \centering
    \includegraphics[width=\columnwidth]{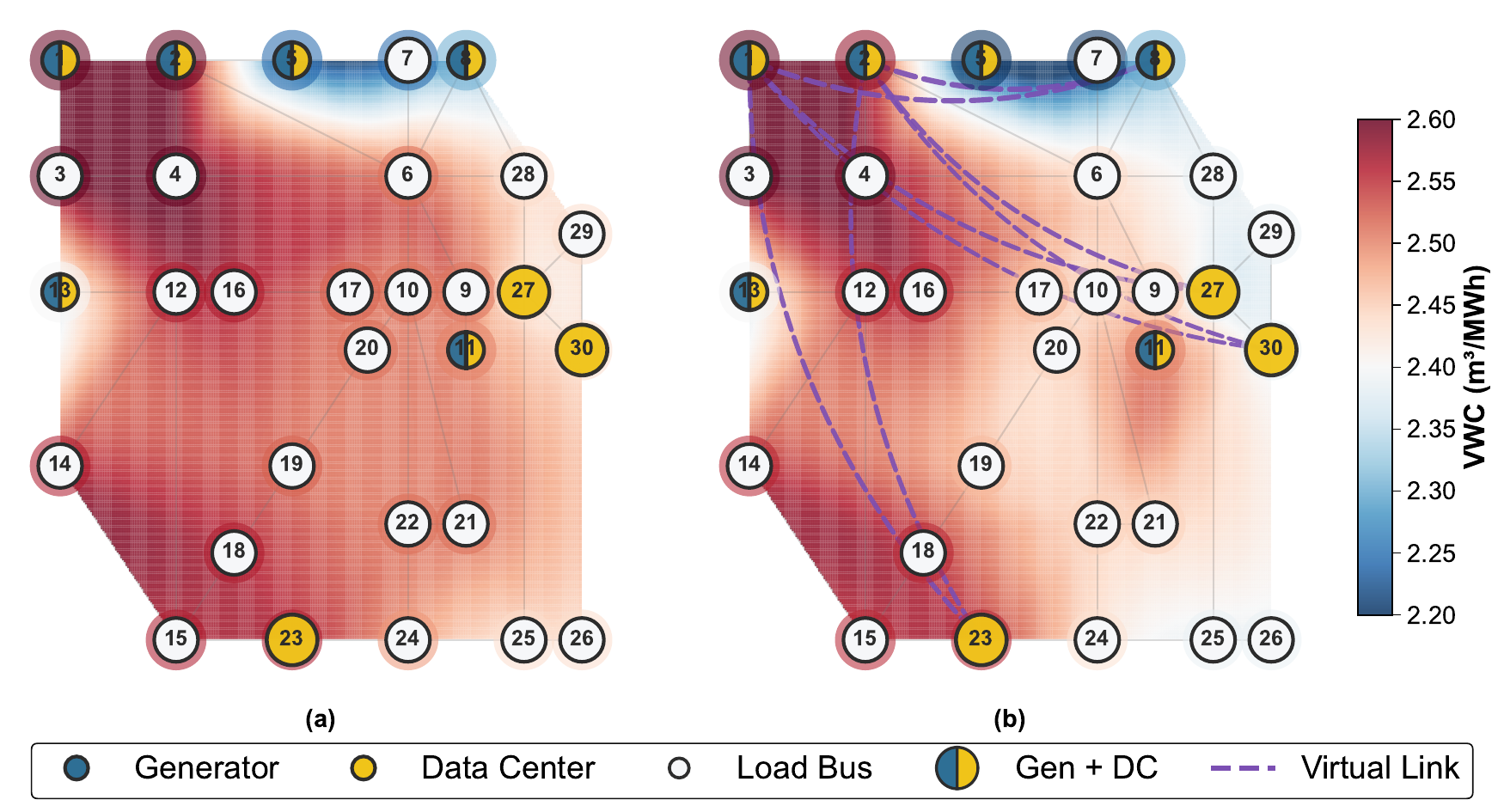}
    \caption{Spatial distribution of virtual water content under (a) Water+VL and (b) water-only configurations.}
    \label{fig:vwc_30bus}
\end{figure}

Fig.~\ref{fig:vwc_30bus} compares the spatial distribution of nodal
virtual water content under the water-only and Water+VL configurations.
Allowing workload migration reduces VWC at several data center buses,
including Buses~6, 20, 21, 25, 26, 27, and~29.
In the water-only case, VWC at these buses ranges from
2.40 to 2.56~m$^3$/MWh, whereas coordinated operation alleviates these
values by redistributing electricity demand away from water-intensive
supply regions.

\begin{figure}[!t]
\centering
\subfloat[$\mathrm{Obj}$ vs.~$W^{\mathrm{budget}}$\label{fig:obj_vs_budget_30}]{%
    \includegraphics[width=0.48\columnwidth]{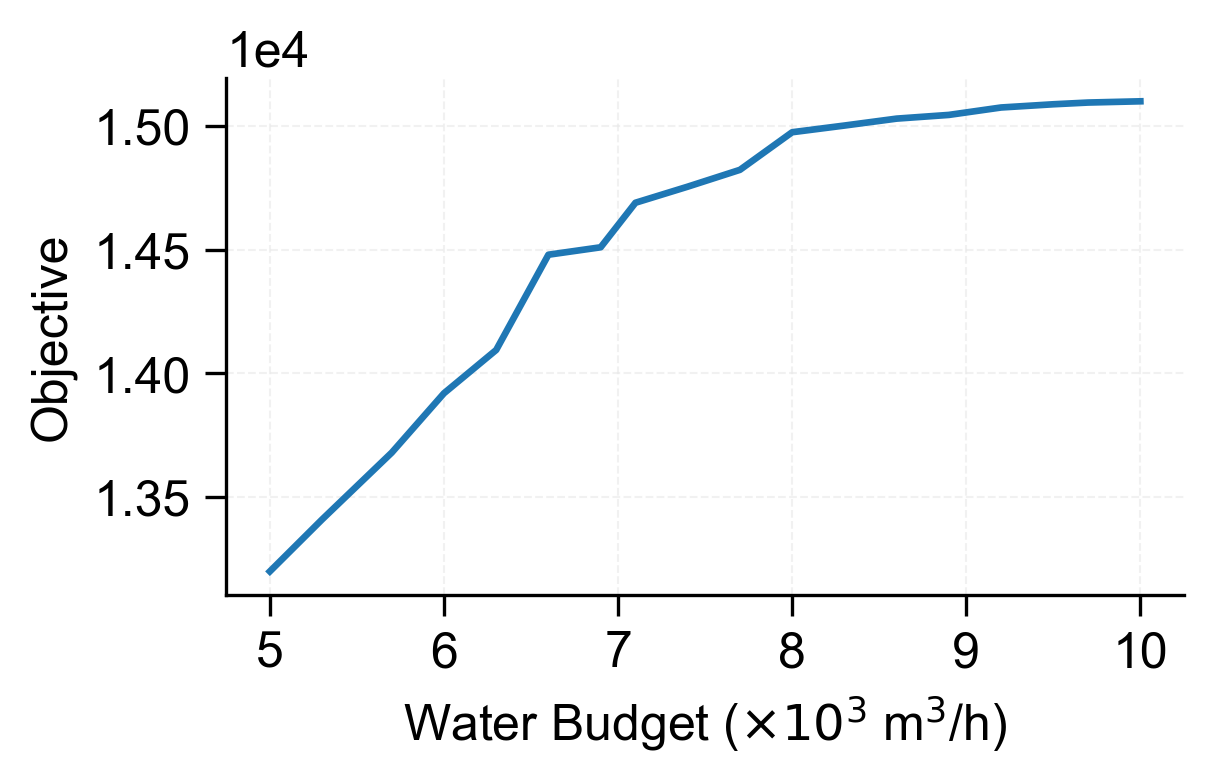}
}\hfill
\subfloat[water withdrawal vs.~$W^{\mathrm{budget}}$\label{fig:water_vs_budget_30}]{%
    \includegraphics[width=0.48\columnwidth]{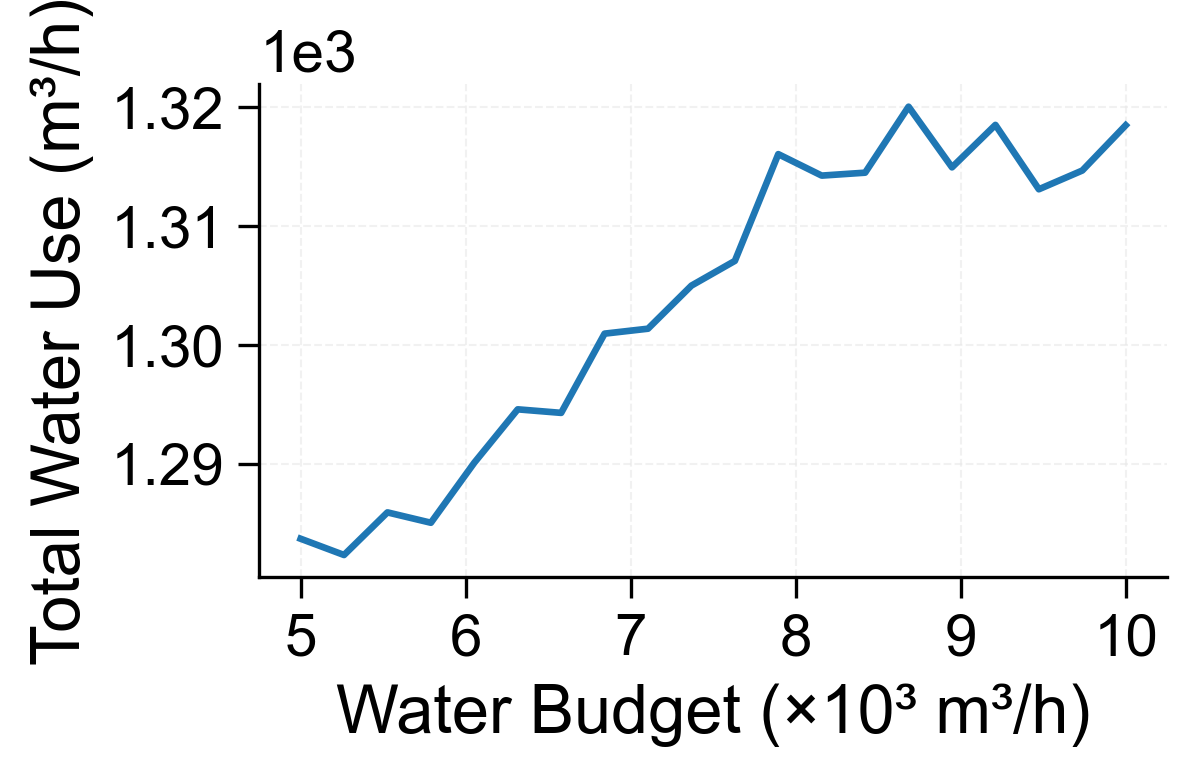}
}\\
\subfloat[water withdrawal vs.~$\lambda_w$\label{fig:water_vs_lambda_30}]{%
    \includegraphics[width=0.48\columnwidth]{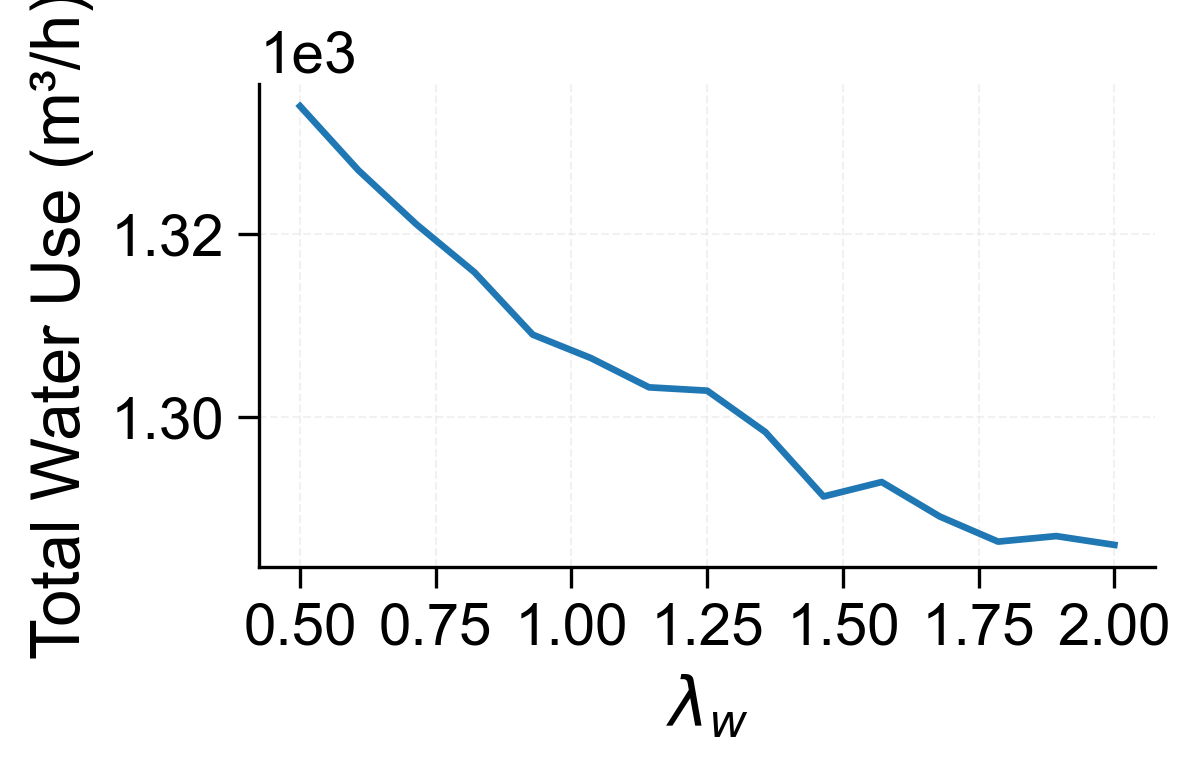}
}\hfill
\subfloat[water withdrawal vs.~$\overline{\Phi}_k$\label{fig:water_vs_vlcap_30}]{%
    \includegraphics[width=0.48\columnwidth]{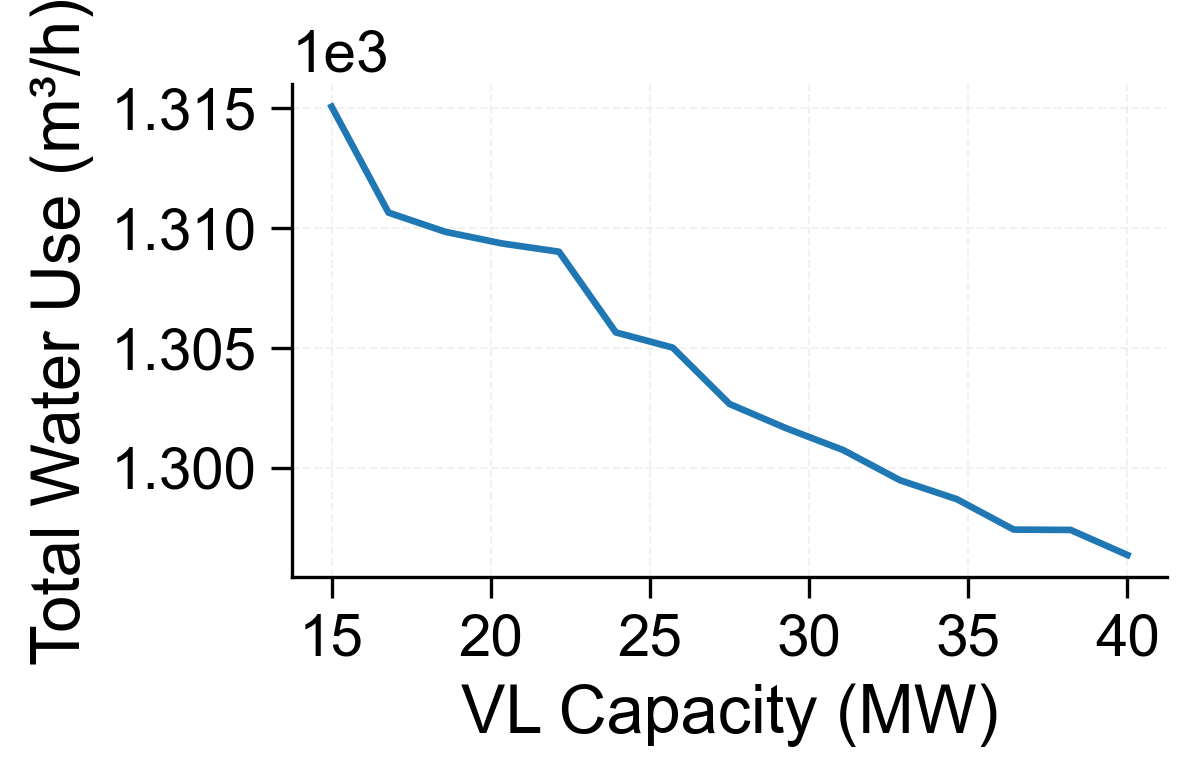}
}
\caption{Sensitivity of system performance to $W^{\mathrm{budget}}$, $\lambda_w$, and $\overline{\Phi}_k$.}
\label{fig:sensitivity_30bus}
\end{figure}

Fig.~\ref{fig:sensitivity_30bus} summarizes system responses.
As $W^{\mathrm{budget}}$ is increased, the objective value rises and
becomes insensitive beyond approximately 9{,}000~m$^3$/h.
Increasing the coefficient $\lambda_w$ in the virtual-water-weighted cost
term~\eqref{eq:cost_water} reduces total water withdrawal, while expanding
$\overline{\Phi}_k$ yields diminishing returns beyond approximately
25~MW.

\subsection{Scalability Analysis on the IEEE 118-Bus System}
\label{sec:ieee118}



\begin{table}[h!]
\centering
\caption{Data center workload redistribution on the IEEE 118-bus system.}
\label{tab:DC_transfer_case118}
\setlength{\tabcolsep}{6pt}
\footnotesize
\begin{tabular}{lcccc}
\toprule
DC & Bus & Baseline (MW) & Optimized (MW) & Change (MW) \\
\midrule
DC1  & 60  & 97.5  & 122.6 & +25.1 \\
DC2  & 78  & 88.9  & 93.1  & +4.2 \\
DC3  & 11  & 82.7  & 84.5  & +1.8 \\
DC4  & 82  & 95.8  & 71.4  & $-24.4$ \\
DC5  & 45  & 92.6  & 69.3  & $-23.3$ \\
DC6  & 88  & 73.2  & 90.7  & +17.5 \\
DC7  & 75  & 76.4  & 78.5  & +2.1 \\
DC8  & 106 & 63.8  & 75.7  & +11.9 \\
DC9  & 95  & 78.6  & 64.2  & $-14.4$ \\
DC10 & 3   & 56.2  & 55.7  & $-0.5$ \\
\midrule
Total & -- & 805.7 & 805.7 & 0.0 \\
\bottomrule
\end{tabular}
\end{table}

Table~\ref{tab:DC_transfer_case118} shows data center workload distributions before and after coordinated operation on the IEEE 118-bus system. Aggregate computing demand remains unchanged at 805.7 MW, while substantial spatial redistribution is observed across data centers. Workloads increase at sites such as DC1 (Bus 60, +25.1 MW) and DC6 (Bus 88, +17.5 MW), and decrease at DC4 (Bus 82, $-24.4$ MW) and DC5 (Bus 45, $-23.3$ MW). The redistribution pattern reflects the framework's ability to shift computing demand toward regions with lower water intensity generation while maintaining system-wide workload balance.

\begin{table}[!t]
\centering
\caption{System-level comparison on the IEEE 118-bus system.}
\label{tab:case118_scenario_comparison}
\renewcommand{\arraystretch}{1.1}
\setlength{\tabcolsep}{4pt}
\scriptsize
\begin{tabular}{lccc}
\toprule
Metric & Optimized & No Coordination & Change (\%) \\
\midrule
Total Generation (MW)
& 4855.04 & 4854.95 & $-0.002$ \\
Load Shedding (MW)
& 3.36 & 3.44 & $-2.41$ \\
Generation Water (m$^{3}$/h)
& 7662.20 & 8060.64 & $-5.20$ \\
Total Objective
& 388{,}492 & 407{,}989 & $-4.78$ \\
\bottomrule
\end{tabular}
\end{table}

Table~\ref{tab:case118_scenario_comparison} compares coordinated and
uncoordinated operation.
Total electricity generation remains essentially unchanged, while
generation-related water withdrawals are reduced by approximately 5.2\%.
Load shedding is slightly lower under coordinated operation.
The reduction in water withdrawal results from spatial reallocation of flexible
computing demand toward regions supplied by lower-intensity generation.

%% file: Case_study/1.2.tex
\subsection{Convergence Analysis and Water Balance Verification}
\label{sec:validation}

The fixed-point iteration is examined on a 5-bus test system to assess
its convergence behavior and to verify consistency between virtual and
physical water accounting.
The compact network size allows detailed inspection of nodal quantities
and transparent visualization of the iterative process.
The system consists of five buses, four generators located at
Buses~1,~2,~4, and~5, and four transmission lines.
Each generator has a maximum capacity of 400~MW, and each transmission
line $(n,m)\in\mathcal{L}$ is subject to a thermal limit of
300~MW.
Generator marginal costs are set to
$(20.0,\,25.0,\,21.0,\,25.0)$~\$/MWh, and water withdrawal 
coefficients are $\kappa_g = (3.80,\,3.20,\,2.60,\,2.30)$~m$^3$/MWh
for the units at Buses~1,~2,~4, and~5, respectively.

To initialize the iteration, the nodal virtual water content
$\mathrm{VWC}_n$ is drawn from a uniform distribution over
$[2.0,\,10.0]$~m$^3$/MWh.
This choice allows convergence to be examined from heterogeneous initial
conditions without bias toward any specific spatial water-use pattern.

\begin{figure}[htbp]
    \centering
    \includegraphics[width=0.85\columnwidth]{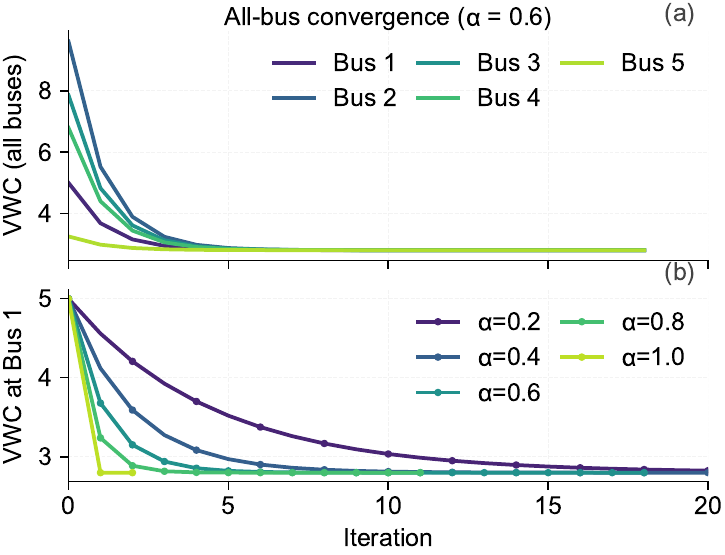}
    \caption{Convergence of nodal virtual water content under the 
    fixed-point iteration: (a)~heterogeneous initial conditions across 
    all buses ($\alpha = 0.6$, $\epsilon = 10^{-6}$); (b)~sensitivity 
    to damping factor~$\alpha$ at Bus~1.}
    \label{fig:vwc_convergence}
\end{figure}

Fig.~\ref{fig:vwc_convergence} illustrates the convergence behavior of 
the fixed-point iteration through two complementary perspectives.
Panel~(a) demonstrates convergence from heterogeneous initial conditions 
across all buses in the 5-bus system.
Starting from arbitrary initial values, the iteration converges within 
6--12 steps, with all buses satisfying the convergence tolerance 
$\epsilon = 10^{-6}$.
Differences in convergence trajectories reflect heterogeneity in network 
connectivity and generation characteristics.
Bus~1, which hosts the most water-intensive generator 
($\kappa_g = 3.80$~m$^3$/MWh), attains the highest equilibrium VWC of 
approximately 3.80~m$^3$/MWh.
In contrast, Bus~4, associated with a low-intensity generator 
($\kappa_g = 2.60$~m$^3$/MWh), converges to the lowest value of 
approximately 2.60~m$^3$/MWh.
The smooth and monotonic evolution across all buses indicates stable 
numerical behavior of the fixed-point mapping under arbitrary 
initialization.

Panel~(b) examines the impact of the damping factor $\alpha$ on 
convergence dynamics at Bus~1.
Five damping factors $\alpha \in \{0.2, 0.4, 0.6, 0.8, 1.0\}$ are 
evaluated, all starting from a fixed initial value of approximately 
5.0~m$^3$/MWh.
The damping factor significantly influences both convergence speed and 
trajectory characteristics.
Smaller values of $\alpha$ yield slower but smoother convergence 
trajectories. 
For instance, $\alpha = 0.2$ requires approximately 18--20 iterations to 
approach the equilibrium value of approximately 2.75~m$^3$/MWh, 
exhibiting gradual monotonic descent without oscillation.
As $\alpha$ increases to 0.4 and 0.6, convergence accelerates 
progressively while maintaining well-damped behavior.
The case $\alpha = 0.6$ reaches equilibrium within 8--10 iterations, 
demonstrating a favorable balance between speed and numerical stability.
At the other extreme, $\alpha = 1.0$ represents undamped updating and 
converges most rapidly in approximately 2--3 iterations, though this 
comes at the expense of trajectory smoothness in the early stages.
The selection of $\alpha = 0.6$ as the default damping parameter 
throughout this study is motivated by its ability to achieve reasonably 
fast convergence without sacrificing numerical stability.
Notably, all damping factors converge to the same equilibrium value of 
approximately 2.75~m$^3$/MWh, confirming that the fixed point is 
independent of the damping schedule and determined solely by the 
underlying power system structure and generator characteristics.

Table~\ref{tab:5bus_water_balance} reports nodal dispatch outcomes and 
the corresponding reconciliation between virtual and physical water 
withdrawals. Total virtual water consumption, computed as
$\sum_{n\in\mathcal{N}} \mathrm{VWC}_n \widetilde{P}_n$,
equals 3148.59~m$^3$/h.
The total physical water withdrawal,
$\sum_{n\in\mathcal{N}} W^{\mathrm{gen}}_n$ with
$W^{\mathrm{gen}}_n$ defined in~\eqref{eq:node_gen_water},
equals 3148.58~m$^3$/h.
The discrepancy of 0.01~m$^3$/h (0.0003\%) is attributable to numerical
rounding.
In this validation case, no data center workload reallocation is applied,
and the effective nodal demand satisfies
$\widetilde{P}_n = P_{d,n}$ for all buses.

\begin{table}[!t]
\centering
\caption{Node-level reconciliation between virtual and physical water withdrawal.}
\label{tab:5bus_water_balance}
\renewcommand{\arraystretch}{1.05}
\setlength{\tabcolsep}{3pt}
\scriptsize
\begin{tabular}{ccccccc}
\toprule
Bus
& $\widetilde{P}_n$
& $P_{g,n}$
& $\mathrm{VWC}_{\mathrm{init}}$
& $\mathrm{VWC}_{\mathrm{final}}$
& $\kappa_g$
& Physical \\
& (MW) & (MW)
& \multicolumn{3}{c}{(m$^3$/MWh)}
& (m$^3$/h) \\
\midrule
1 & 200.00 & 400.00 & 4.371 & 3.800 & 3.80 & 1520.00 \\
2 & 220.00 &  66.19 & 9.556 & 3.651 & 3.20 &  211.82 \\
3 & 290.00 &   0.00 & 7.588 & 2.753 & --   &    0.00 \\
4 & 170.00 & 400.00 & 6.388 & 2.600 & 2.60 & 1040.00 \\
5 & 150.00 & 163.81 & 2.404 & 2.300 & 2.30 &  376.76 \\
\midrule
Sum & 1030.00 & 1030.00 & -- & -- & -- & 3148.58 \\
\midrule
\multicolumn{6}{l}{Virtual: $\sum_n \mathrm{VWC}_n \cdot \widetilde{P}_n$ (m$^3$/h)}
& 3148.59 \\
\bottomrule
\end{tabular}
\end{table}

At the nodal level, Bus~1 generates 400~MW using 1520~m$^3$/h of water but
serves only 200~MW of local demand, resulting in exports of electricity
and its associated embedded water.
Bus~3, which has no local generation, imports all of its electricity and
exhibits a virtual water footprint determined entirely by upstream
sources.
These patterns illustrate how nodal virtual water content captures the
spatial redistribution of water embodied in electricity flows.

Overall, the fixed-point solution preserves consistency between virtual
and physical water accounting, such that the aggregate virtual water
embedded in electricity consumption matches total physical withdrawals
within numerical tolerance.
When nodes host generation with negligible water withdrawal, exported
electricity carries correspondingly low embedded water, reducing
system-wide water intensity.

%% file: Case_study/1.3.tex
\subsection{Coordinated Operation with Water Constraints}
\label{sec:vl_coordination}

This section evaluates coordinated operation on the 5-bus system
with water scarcity constraints and workload migration flexibility.
Unlike the validation case in Section~\ref{sec:validation}, this 
configuration includes water withdrawal limits and enables spatial 
workload redistribution.
For clarity, the time index $t$ is omitted.

Generators are located at Buses~1, 2, 4, and~5, with water withdrawal 
coefficients $\kappa_g = 2.8$, $0.6$, $1.2$, and $2.3$~m$^3$/MWh, 
respectively.
The baseline nodal electricity demands are
$P_{d,n} = 55.0$, $55.0$, $62.0$, $60.0$, and $66.0$~MW at Buses~1 through~5.
The water scarcity weights $s_n$ in
\eqref{eq:scarcity_weighted_budget} are set to
$1.5$, $2.0$, $1.0$, $0.5$, and $2.0$, respectively.
The system level scarcity-weighted water constraint
\eqref{eq:scarcity_weighted_budget} is enforced with
$W^{\mathrm{budget}} = 5{,}000$~m$^3$/h.
In addition, nodal water withdrawal capacities are constrained by
local water availability, with maximum nodal withdrawals set to
$\overline{W}^{\mathrm{gen}}_{n} =
800$, $500$, $800$, $1200$, and $600$~m$^3$/h at Buses~1 through~5.
Each transmission line $(n,m)\in\mathcal{L}$ is subject to a flow limit
$\overline{F}_{nm} = 250$~MW.
Workload migration is enabled with symmetric link capacities
$\overline{\Phi}_k = 20$~MW.
Two configurations are compared.
The baseline case excludes water constraints and workload migration. The coordinated case (Water+VL) enforces the system-level water 
constraint~\eqref{eq:scarcity_weighted_budget} along with nodal 
water withdrawal limits, and enables workload migration.

In the baseline dispatch, Buses~1 and~4 operate at their maximum limits,
while Bus~2 produces 81.79~MW and Bus~5 produces 283.21~MW.
Under coordinated operation, generation shifts toward the water-efficient
unit at Bus~2, which increases to 130.00~MW, while generation at Bus~5
decreases to 235.00~MW.
Buses~1 and~4 remain capacity-constrained.

\begin{table}[!t]
\centering
\caption{Node-level water balance: baseline vs.\ Water+VL operation.}
\label{tab:water_balance_comparison}
\renewcommand{\arraystretch}{1.05}
\setlength{\tabcolsep}{2pt}
\scriptsize
\begin{tabular}{ccccccc}
\toprule
\multirow{2}{*}{Bus} &
\multicolumn{3}{c}{Baseline (No Coordination)} &
\multicolumn{3}{c}{Water+VL}\\
\cmidrule(lr){2-4}\cmidrule(lr){5-7}
 & $P_{g,n}$ & $\widetilde{P}_n$ & $\mathrm{VWC}$ &
   $P_{g,n}$ & $\widetilde{P}_n$ & $\mathrm{VWC}$\\
 & (MW) & (MW) & (m$^3$/MWh) &
   (MW) & (MW) & (m$^3$/MWh)\\
\midrule
1 & 400.00 & 250.00 & 2.800 & 400.00 & 217.70 & 3.046\\
2 &  81.79 & 190.00 & 2.024 & 130.00 & 190.00 & 1.762\\
3 &   0.00 & 240.00 & 1.473 &   0.00 & 252.30 & 1.360\\
4 & 400.00 & 230.00 & 1.200 & 400.00 & 230.00 & 1.200\\
5 & 283.21 & 255.00 & 2.300 & 235.00 & 275.00 & 2.187\\
\midrule
Total & 1165.00 & 1165.00 & -- & 1165.00 & 1165.00 & --\\
\midrule
\multicolumn{7}{l}{\scriptsize System water withdrawal (m$^3$/h): 2300.46 $\rightarrow$ 2218.50 ($-3.6$\%)}\\
\bottomrule
\end{tabular}
\end{table}

Table~\ref{tab:water_balance_comparison} shows that nodal virtual water
content varies across the network due to differences in local generation
and imported electricity.
Under coordinated operation, workload migration alters the effective
demand pattern, which shifts dispatch toward the low-$\kappa_g$
generator at Bus~2 and reduces total water withdrawal by 3.6\%.
At the nodal level, $\mathrm{VWC}$ adjusts consistently with the induced
changes in physical power flows.

%% file: appendix.tex
\appendix
\section{Derivation of Gradients for the Optimization Layer}
\label{app:gradient_derivation}

This section derives the backward pass of the optimization layer by implicit differentiation of the KKT conditions.

The optimization layer solves the quadratic program
\begin{equation}
z^* = \arg\min_{z} \frac{1}{2} z^{T} Q z + q^{T} z 
\quad \text{s.t. } Az = b,\; Gz \le h .
\end{equation}
Let $(z^*, \nu^*, \lambda^*)$ denote the primal and dual optimal variables. The KKT conditions are
\begin{equation}
\begin{aligned}
Q z^* + q + A^{T} \nu^* + G^{T} \lambda^* &= 0, \\
A z^* - b &= 0, \\
D(\lambda^*)(G z^* - h) &= 0 .
\end{aligned}
\end{equation}

Taking the total differential of the KKT system at the optimum yields a linear system in $(dz, d\lambda, d\nu)$:
\begin{equation}
\begin{aligned}
&\begin{bmatrix}
Q & G^{T} & A^{T} \\
D(\lambda^*)G & D(Gz^* - h) & 0 \\
A & 0 & 0
\end{bmatrix}
\begin{bmatrix}
dz \\ d\lambda \\ d\nu
\end{bmatrix} \\
&= -
\begin{bmatrix}
dQ z^* + dq + dG^{T} \lambda^* + dA^{T} \nu^* \\
D(\lambda^*) dG z^* - D(\lambda^*) dh \\
dA z^* - db
\end{bmatrix}.
\end{aligned}
\end{equation}

In the backward pass, given the loss gradient $\partial l / \partial z^*$, we solve for the adjoint variables $(d_z, d_\lambda, d_\nu)$ using the transposed KKT system:
\begin{equation}
\begin{bmatrix} d_{z} \\ d_{\lambda} \\ d_{\nu} \end{bmatrix}
=
-\begin{bmatrix}
Q & G^{T} D(\lambda^*) & A^{T} \\
G & D(G z^* - h) & 0 \\
A & 0 & 0
\end{bmatrix}^{-1}
\begin{bmatrix}
(\frac{\partial l}{\partial z^*})^{T} \\ 0 \\ 0
\end{bmatrix}.
\end{equation}

The gradients with respect to the problem parameters are then given by
\begin{equation}
\begin{aligned}
\nabla_{Q} l &= \tfrac{1}{2} (d_{z} z^{*T} + z^* d_{z}^{T}), 
& \nabla_{q} l &= d_{z}, \\
\nabla_{A} l &= d_{\nu} z^{*T} + \nu^* d_{z}^{T}, 
& \nabla_{b} l &= - d_{\nu}, \\
\nabla_{G} l &= D(\lambda^*) d_{\lambda} z^{*T} + \lambda^* d_{z}^{T}, 
& \nabla_{h} l &= - D(\lambda^*) d_{\lambda}.
\end{aligned}
\end{equation}